# ChatSOS: Vector Database Augmented Generative Question Answering Assistant in Safety Engineering

Haiyang Tang, Dongping Chen[*], Qingzhao Chu[*]

*School of Mechatronic Engineering, Beijing Institute of Technology, Beijing, China*

**Abstract:** With the rapid advancement of natural language processing technologies, generative artificial intelligence techniques, represented by large language models (LLMs), are gaining increasing prominence and demonstrating significant potential for applications in safety engineering. However, fundamental LLMs face constraints such as limited training data coverage and unreliable responses. This study develops a vector database from 117 explosion accident reports in China spanning 2013 to 2023, employing techniques such as corpus segmenting and vector embedding. By utilizing the vector database, which outperforms the relational database in information retrieval quality, we provide LLMs with richer, more relevant knowledge. Comparative analysis of LLMs demonstrates that ChatSOS significantly enhances reliability, accuracy, and comprehensiveness, improves adaptability and clarification of responses. These results illustrate the effectiveness of supplementing LLMs with an external database, highlighting their potential to handle professional queries in safety engineering and laying a foundation for broader applications.

## 1. Introduction

With the development of natural language processing (NLP) technology and the rapid increase in computing capability, large language models (LLMs) based on the Transformer architecture have demonstrated high levels of intelligence. Notably, a series of LLMs were introduced by leading companies such as OpenAI, Meta, and Google, including GPTs (Achiam et al., 2023; Brown et al., 2020; Radford et al., 2018; Radford et al., 2019), Llama (Roziere et al., 2023; Touvron et al., 2023a; Touvron et al., 2023b), Bard, and Gemini. LLMs have made significant breakthroughs in diverse research fields, ranging from software engineering to biochemical research based on natural language dialogue.

On March 14, 2023, OpenAI officially released the most powerful LLM to date, e.g., GPT-4. This model has demonstrated excellent performance in SAT and BAR exams, LeetCode challenges, and context extraction from images (Achiam et al., 2023). Relying on the high intelligence and powerful generative capabilities of such models, researchers are now exploring the potential of LLMs across numerous industries, seeking to unlock new practical applications. Kasneci et al. (Kasneci et al., 2023) proposed a method using generative language models to enhance education and research experiences.

---
[*] Corresponding authors. Tel: +86 13167557765.
e-mail: dc516@bit.edu.cn (Dongping Chen); chuqz@bit.edu.cn (Qingzhao Chu).

Wang et al. (Wang et al., 2024) introduced the innovative concept of LLM system engineering (LLM-SE), incorporating methodologies from knowledge engineering and quality engineering, and developed a clinically oriented risk prediction system. Giorgio Valentini et al. (Valentini et al., 2023) highlighted the emergence of neural network models that can predict protein properties and create proteins with novel functionalities. This paper demonstrates the effectiveness of multimodal protein language models (PLM) in addressing protein modelling challenges and enhancing analytical model prediction mechanisms. Recently, researchers have been dedicated to integrating agents with LLMs. Agents possess perceptual, control, and action capabilities, enabling them to parse user demands, formulate workflows, and execute operations (Xi et al., 2023). This integration facilitates cross-program interaction capabilities, which are challenging to achieve solely by relying on LLMs themselves. Vaghefi et al. (Vaghefi et al., 2023) developed a model agent that integrates diverse information sources to mitigate the language accuracy and credibility limitations in LLMs, particularly in addressing climate-related issues where relying solely on original model weights proves insufficient. Moreover, as the application of LLM technology expands across diverse industries, evaluating its effectiveness in the field of safety engineering has become increasingly imperative.

Safety engineering, as an interdisciplinary field, requires the comprehensive application of various scientific principles and methods to prevent, reduce, or respond to safety risks and accidents, safeguarding human life, property, and the environment in a comprehensive, sustainable, and resilient manner (Ouyang et al., 2018). Traditional analysis methods for safety accident includes Integrated inherent safety index (Khan and Amyotte, 2004), Root Cause Analysis (Andersen and Fagerhaug, 2006), Event Tree Analysis, Risk Matrix Analysis, etc. These analysis methods are often used in combination to understand the causes, possibilities, and impacts of accidents. In the era of artificial intelligence (AI) today, state-of-the-art techniques, such as LLMs, can be introduced to assist in the analysis of safety risks and accidents. However, the direct application of LLMs in the above scenarios may have inherent limitations due to the lack of professional knowledge and the model hallucinations (Church and Yue, 2023; Oviedo-Trespalacios et al., 2023).

To address the above shortcomings, the integration of professional databases with LLMs offers a promising approach for enhancing accuracy in professional contexts. Databases occupy a pivotal position in practical engineering applications, and their ubiquity across diverse industries is testament to their significance. Relational database management systems (RDBMSs) have been designed to manage and query structured data, exceling in precise searches based on exact word matching. Their ability to store information in a tabular format ensures data consistency and integrity, which is crucial for maintaining the integrity of complex systems. However, the strict requirements for data formatting in RDBMS can be cumbersome, particularly during the preprocessing stage. Furthermore, their limitations in capturing semantic similarities become evident when dealing with NLP tasks in the realm of information retrieval. This gap highlights the need for innovative solutions that can offer good solutions for both structured and unstructured data. With the rapid advancement of NLP technology, vector databases have emerged as promising alternatives. Unlike traditional relational databases, vector databases are designed to store various types of unstructured data. They leverage pretrained encoding models to embed text data into a high-dimensional vector space, where semantic relationships between text contents are preserved. The corresponding data retrieval method relies on calculating vector similarity to identify the most relevant text sequences. Widely used vector database management systems include Milvus (Wang et al., 2021) and Facebook AI similarity search (Faiss) (Jégou et al., 2022; Johnson et al., 2019). Vector databases have significant advantages in terms of data scale, retrieval methods, and data flexibility (Seidl and Kriegel, 1997). These advantages make them highly compatible with LLMs that require massive amounts of data.

In this work, a vector database-augmented generative question answering assistant in safety engineering is introduced by combining an LLM and a vector database. This assistant avoids model hallucinations by extracting content from a local professional database as supplementary knowledge. This process improves the accuracy of model responses, ensuring the reliability of large models. To validate the performance of the assistant in improving the deficiencies of basic LLMs, this work conducts a series of evaluations in various scenarios. Furthermore, this work developed a scoring criterion for the model, aiming to achieve more practical and valuable assessment of an AI-driven question-answering assistant.

## 2. Method
### 2.1 ChatSOS Framework

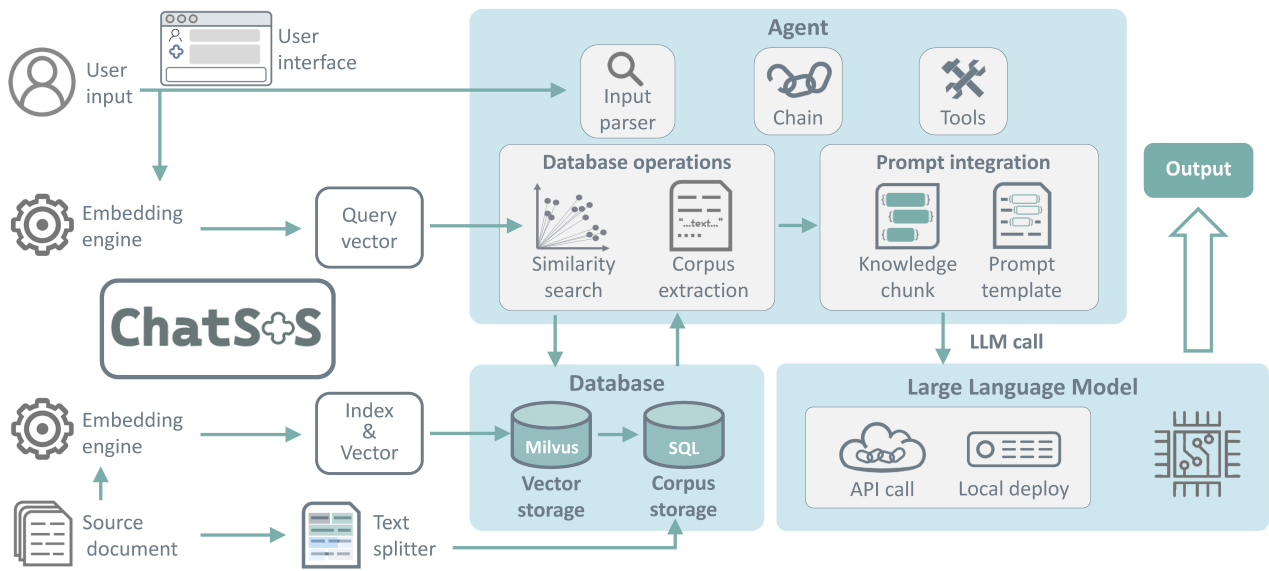

**Fig. 1.** ChatSOS framework.

In this work, we present an AI-driven question-answering assistant, e.g., ChatSOS (<u>Chat</u> <u>S</u>ecurity <u>O</u>racle<u>s</u>), for safety engineering. ChatSOS comprises three primary components: a database, an LLM, and an agent. The database incorporates relational and vector databases, collectively feeding relevant text into the LLM to facilitate answers. The LLM, trained on extensive textual data, possesses robust natural language comprehension and generation capabilities, enabling it to fulfil specific tasks or requirements. The agent, serving as a crucial component within the Langchain (Chase, 2022), is driven by the language model and prompts, governs the responsibility of determining subsequent actions. The agent is designed to augment the capabilities of the LLM, addressing the challenges of precise matching between users and the tool. By harnessing these components, ChatSOS offers a comprehensive and efficient solution for addressing user queries.

## 2.2 Database

In any discipline, professional definitions and basic concepts can be saved as text or as data that can be converted to text, which serves as the foundation for a question-answering assistant. The search methodology of the traditional RDBMS accurately retrieves results containing exact keywords but fails to return an outcome with similar concepts, deviating from human thinking methods. Human cognition often involves recognizing and comprehending entities through the construction of relationship networks, necessitating heightened generalizability when retrieving information from a database. While the traditional RDBMS is well suited for storing and analysing structured data, a vector database is more suitable for accommodating unstructured data. In ChatSOS, the vector database empowered by Milvus (Wang et al., 2021) is used to manage the data storage and vector similarity retrieval functions, utilizing a relational database, i.e., MySQL, to store vector IDs and the corresponding text chunks. The database used in this study consists of 117 documents. During the processing of these documents, the textual corpus is segmented by a tokenizer, dividing it into text chunks of ~300 tokens. Each text chunk includes an overlap of 50 tokens with its neighboring chunks. Here, a word or a symbol is treated as one token. Depending on the size of the document, the number of text chunks varies. The text chunks then undergo vector embedding by the BAAI general embedding (BGE) model, mapping them into a vector space and storing in a relational database with unique vector IDs. Text embedding techniques, such as word2vec (Mikolov et al., 2013) and BERT (Devlin et al., 2018), embed text content into high-dimensional vector space based on machine learning methods. Embedded vectors generally contain more information than a standard table, correlating with an increased vector dimension. Mathematical measures are applied to quantify the similarity between the user's query vector and the content vectors in the database. The vector database returns the top-$k$ results by similarity ranking, providing an effective means of information retrieval. There are two commonly used methods for calculating the Euclidean distance and cosine similarity (Salton et al., 1975):

$$Euclidean\ Distance(A,B) = \sqrt{\sum_{i=1}^{n}(x_i - y_i)^2}, \tag{1}$$

$$Cosine\ Similarity(A,B) = \frac{A \cdot B}{\|A\| \cdot \|B\|} = \frac{\sum_{i=1}^{n}(x_i \times y_i)}{\sqrt{\sum_{i=1}^{n}x_i^2} \times \sqrt{\sum_{i=1}^{n}y_i^2}}, \tag{2}$$

where $A(x_1, x_2, \dots, x_n)$ and $B(y_1, y_2, \dots, y_n)$ represent two $n$-dimensional vectors. The Euclidean distance between them can be understood as the "straight-line" distance in an $n$-dimensional space, also known as the L2 score. This metric focuses on the absolute positions of the points. Cosine Similarity measures the similarity in direction between two vectors, which is independent of the magnitude of the vectors.

During actual question-and-answer interactions, the user query is first embedded into the vector space using the BGE model. Subsequently, the system conducts vector similarity calculations in a vector database to identify the top 10 vectors with the highest similarity scores along with their corresponding IDs. These IDs are then used to retrieve the corresponding text chunks from the original corpus stored in the relational database. These text chunks are provided as supplementary knowledge to the agent and integrated into the prompt template for subsequent operations. In this process, the use of a verified local database offers a reliable source of information for the LLM. Additionally, by employing vector similarity calculations, it is possible to find information with better relevance in the semantic space based on similarity scores.

The ChatSOS includes a database that comprises a collection of 117 accident investigation reports focusing on combustion and explosion accidents in China. Among these reports, two are issued by national authorities, while the remaining reports cover cases in various provinces, spanning from 2013 to 2023. Figure 2 illustrates the spatial, industrial, and temporal distributions of the accident investigation reports in this study. Figure 2a illustrates the spatial distribution of the reports categorized by province. The five provinces with the most accident reports are Shandong, Liaoning, Jiangsu, Henan, and Hunan, with 13, 9, 8, 7, and 7 reports, respectively. It is noteworthy that the dataset exhibits a greater incidence of accidents in the eastern region than in the western region. Figure 2b shows the classification of accidents based on the industries involved, revealing a predominant occurrence of explosion accidents within the mining and biomedical and chemistry domains, constituting more than 50% of the dataset. Figure 3c provides a temporal perspective of the collected accident reports. The maximum number of reports is 19, taken from 2019, while we only have 5 and 1 reports for 2022 and 2023, respectively. In addition to the above three years, the number of reports fell in the range of 8-14.

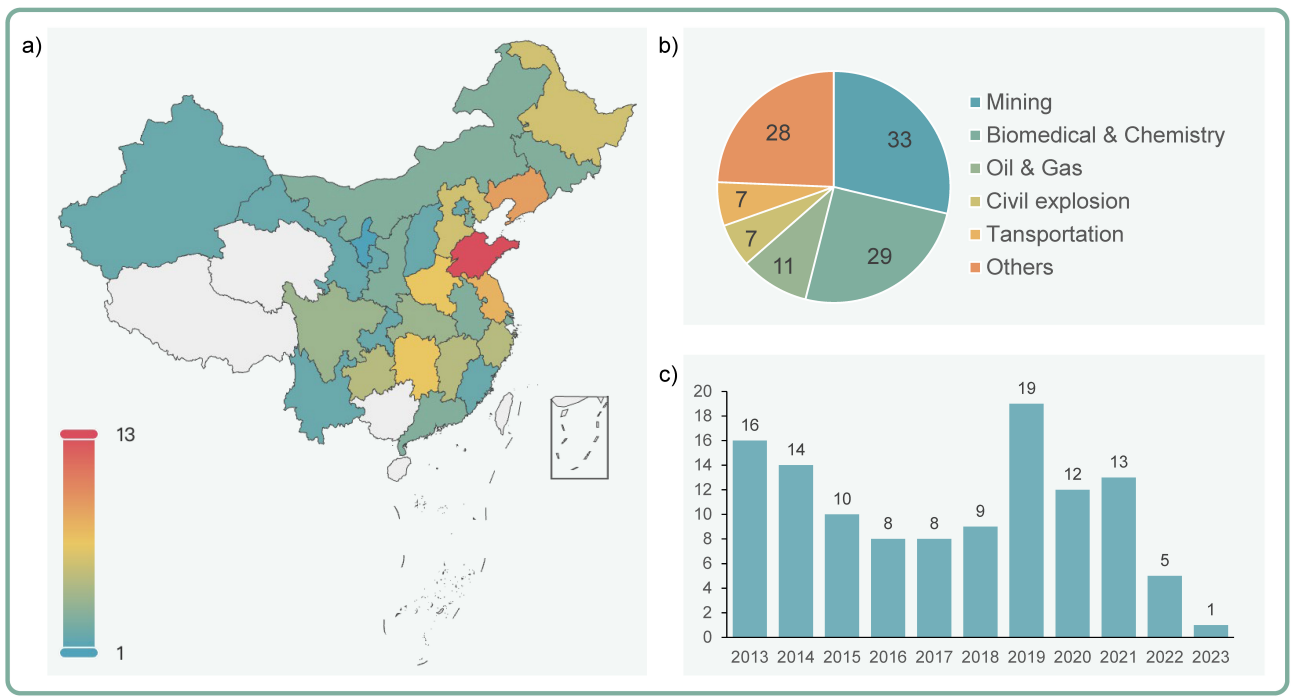

**Fig. 2.** Accident investigation database in this work, soring by (a) province in China, (b) industries, and (c) years.

### 2.3 Large Language Models

This study conducts comparative experiments using three LLMs: ChatGPT by OpenAI, ERNIE Bot by Baidu, and ChatSOS proposed in this paper. In these experiments, ChatSOS utilized APIs to connect with the GPT-3.5-Turbo model for testing, and the output results are compared with those of ChatGPT to verify the effectiveness of the vector database-augmented approach in the field of safety engineering. The inclusion of the ERNIE Bot allows for identifying the differences among available

LLMs, while also mitigating the potential randomness of results from ChatGPT.

The present study delves into various prompt templates and scenario presets to enhance the responses of the LLM (Jha et al., 2023). In ChatSOS, a series of prompt templates tailored for different task types are configured to provide supplementary knowledge and standardize model responses. These templates improve the performance of LLMs according to the requirements of the task, ensuring precise and relevant outputs. Our prompt augmenting approach implements highly relevant knowledge injection based on the task type of the user input. Figure 3a is a prompt template in ChatSOS. We supplement LLMs with multiple Q&A samples to enhance reliability. The model is instructed to provide answers based solely on given reference materials. If the provided information is insufficient to draw conclusions, the LLM is encouraged to output "Insufficient information to answer". This strategy aims to enhance the model's understanding and adaptation to user-specific requirements, thereby improving its overall performance and reliability. Figure 3b presents the full prompt for an actual question-and-answer interaction. The user enquires the ChatSOS for "Analysis of responsibility for the major explosion accident at Shandong Sifang Installation Engineering Co., Ltd.". The prompt defines the roles of the LLM and the user, integrating preset scenarios and custom prompt templates. Notably, database retrieval results are incorporated as supplementary knowledge alongside user queries within the specific section of the template.

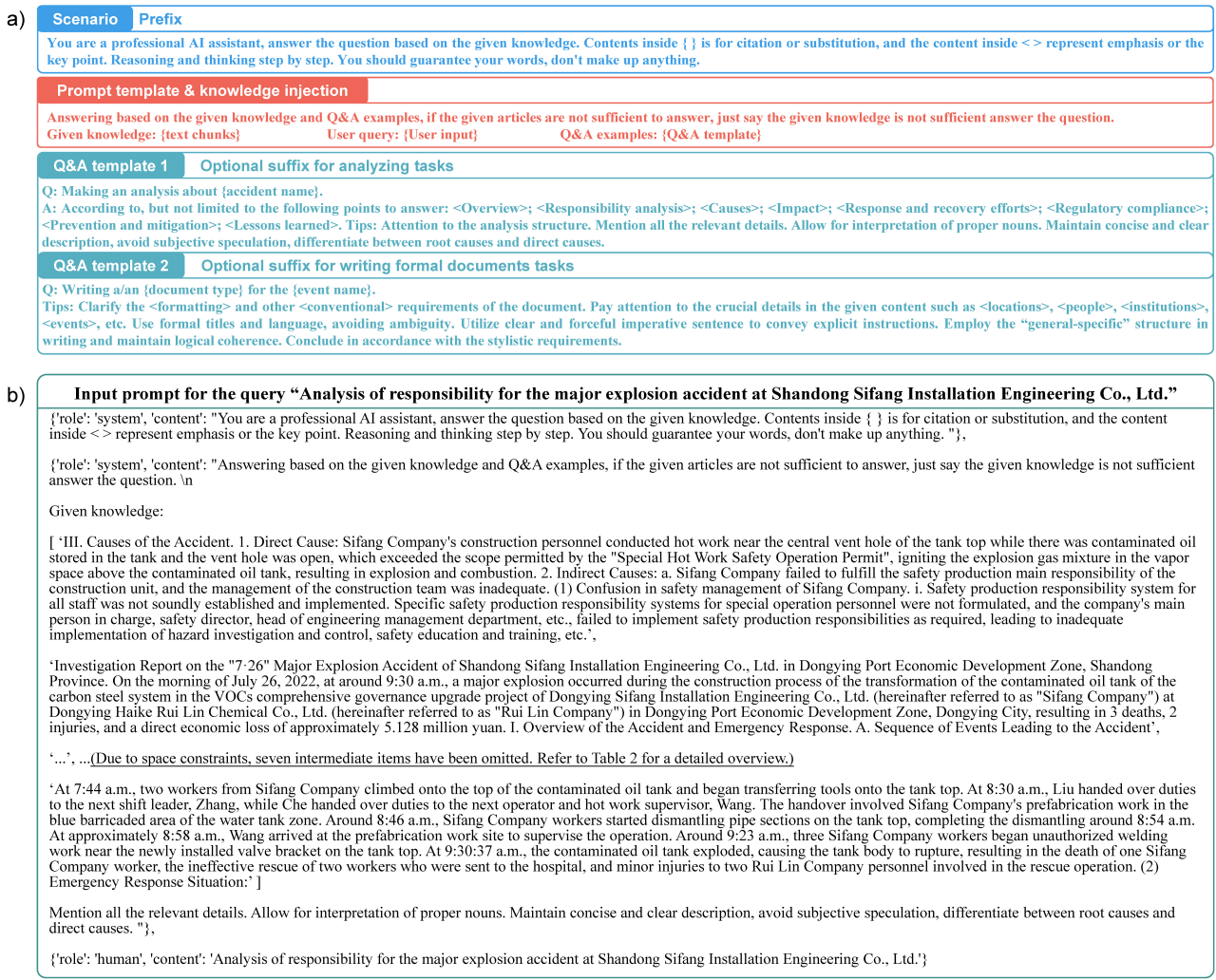

**Fig. 3.** (a) A prompt template in ChatSOS, and (b) an input prompt in the Q&A interaction.

## 2.4 Agent

The agent serves as the central component within the LangChain framework(Chase, 2022), playing a pivotal role in enhancing the capabilities of LLMs. This module functions as a coprocessor, which is responsible for receiving user inputs, executing actions, and generating corresponding outputs, thereby augmenting the functionality and adaptability of LLMs. The core functionality of agent is underpinned by its ability to integrate various specialized modules, including information retrieval, user interaction, and specific toolkits. This integration facilitates effective interaction between LLMs and the external database, enhancing the intelligence and adaptability of LLMs in addressing real-world problems. Figure 4 demonstrates the process of handling Q&A tasks in ChatSOS using a local database as a knowledge supplement. During the Q&A process, the agent parses the input to understand the user's intent and creates a work chain tailored to the specific work. This work chain includes selecting appropriate tools, matching prompt templates, and integrating results from various sources. The agent then leverages the vector database to perform similarity searches, retrieving text chunks from the preset top-$k$ ranking. The text chunks further served as supplementary information in the prompt. By integrating the results from the vector database and other tools, the agent builds a final prompt that is appropriate for the user's query and optimized for input into the LLM. The LLM then processes this prompt and generates a response that is informed of the user's queries.

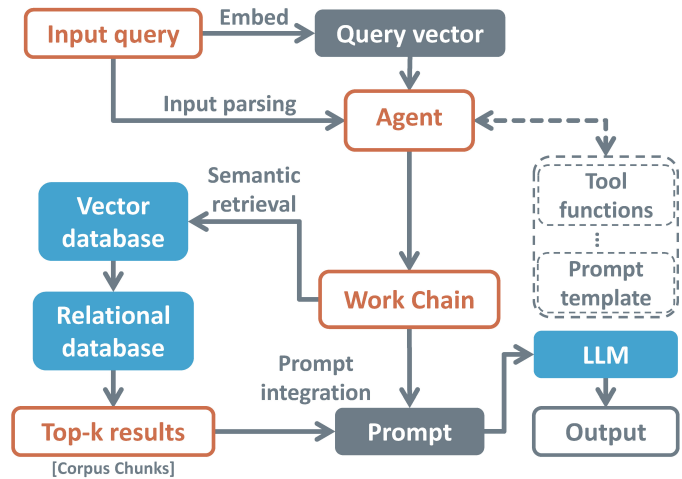

**Fig. 4.** Workflow of Q&A tasks in ChatSOS.

## 3. Results and discussion

### 3.1 Database retrieval evaluation

To assess the accuracy of processing natural language in database retrieval, this study devised three sets of retrieval tasks, e.g., Group A, B and C, across different scenarios, comparing the performance of relational database and vector database. In Table 1, Group A tasks focus on searching for individual entity names and technical terms to evaluate the generalization capabilities of the databases. Group B tasks retrieve short phrases directly from source documents, evaluating the accuracy of the retrieval and the ability to retrieve associated text chunks. Group C tasks perform retrievals based on detailed user queries about a particular accident event, aiming to test the comprehensive performance of the database for practical Q&A queries.

**Table 1**
Designed terms for database retrieval evaluation

| Term | Content[a] |
|------|-----------|
| A-1  | Gas explosion |
| A-2  | Hongtutian coal mine |
| A-3  | Nanyang Fireworks Factory |
| B-1  | Disposal Situation of Jincheng Expressway Limited Liability Company |
| B-2  | Dongfang Coal Mine, Xiejiaji District, Huainan City, Anhui Province |
| B-3  | During the operation of the electric tar catcher, the current and voltage are unstable, and the discharging is severe |
| C-1  | A brief summary of the explosion accident at Shanghai Qingpu Hengchang Logistics Co., Ltd. |
| C-2  | Analysis the responsibility for the explosion accident at Shandong Sifang Installation Engineering Co., Ltd. |
| C-3  | The reasons for the gas explosion accident at Hongxin Coal Industry Company in Wangcang on July 28$^{th}$ |

[a] The search terms used for retrieval are in Chinese. Here, the translated terms/phrases are shown.

In the retrieval tasks, we use the number of returned entries (i.e., text chunks) as the effective count for the relational database, as it returns entries that exactly match the search terms. However, the top-$k$ method in the vector database is considered for evaluating the performance of the vector database, because the vector database ranks all text chunks based on L2 scores. Moreover, LLMs utilize a sliding window mechanism during text generation, which is limited by the parameters of the context window. In this work, we use the GPT-3.5-turbo-0125 model as the LLM. It supports 16k tokens for the context window, suggesting that we cannot support a context size exceeding this limit. Otherwise, we should potentially observe a loss of information or breaks in the text continuity. To ensure the concentration of relevant information in the supplementary information provided to the LLM, it is crucial to determine an appropriate $k$ value. The maximum $k$ value can be estimated as 50 considering the size of each text chunk (i.e., ~300 tokens). We select the top-50 method as the reference to assess the performance of the vector database, and manually analyze the relevance of the text chunks related to the search terms. Two other methods, i.e., top-10 and top-20, are considered to evaluate the accuracy of the retrieved results (Figure 5).

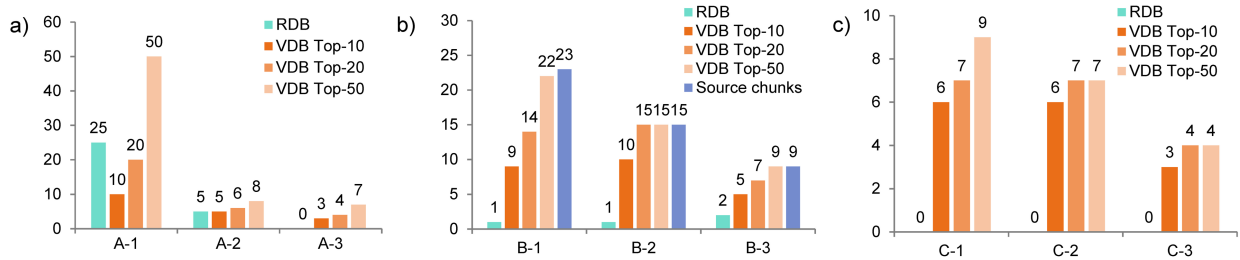

**Fig. 5.** Number of correct entries of relational database (RDB) and vector databases (VDB). (a) Group-A: keyword retrievals; (b) Group-B: retrievals of original sentences; (c) Group-C: retrievals of natural language questions. The design terms for database retrieval are included in Table 1.

**Table 2**
Retrieval performance of the vector database with the top-$k$ method.

| Term | PRE[a] (%) | | | PRE-50[b] (%) | | | Recall rate[c] (%) | | |
|---|---|---|---|---|---|---|---|---|---|
| $k$ | 10 | 20 | 50 | 10 | 20 | 50 | 10 | 20 | 50 |
| A-1 | 100 | 100 | 100 | 20 | 40 | 100 | - | - | - |
| A-2 | 50 | 30 | 16 | 62.5 | 75 | 100 | - | - | - |
| A-3 | 30 | 20 | 14 | 42.9 | 57.1 | 100 | - | - | - |
| B-1 | 90 | 70 | 44 | 40.9 | 63.6 | 100 | 75 | 60.9 | 95.7 |
| B-2 | 100 | 75 | 30 | 66.7 | 100 | 100 | 66.7 | 100 | 100 |
| B-3 | 50 | 35 | 18 | 55.6 | 77.8 | 100 | 55.6 | 77.8 | 100 |
| C-1 | 60 | 35 | 18 | 66.7 | 77.8 | 100 | - | - | - |
| C-2 | 60 | 35 | 14 | 85.7 | 100 | 100 | - | - | - |
| C-3 | 30 | 20 | 8 | 75 | 100 | 100 | - | - | - |

[a] Proportion of relevant entries among the top-$k$ results.
[b] Proportion of relevant top-$k$ entries to the top-50 relevant results.
[c] The recall rate is determined by the number of top-$k$ results from the source chunks.

For Group A, simple terms, such as "Gas explosion", "Hongtutian coal mine" and "Nanyang Fireworks Factory", are considered. The first two terms are taken directly from source documents, but the last term is an abbreviated business name that does not exist in such a form. The relational database returns 25 and 5 results for the first two terms, respectively. However, this approach fails to return any results for the last one, as relational database can only extract correct results that exactly match the search terms. Based on the vector database, 10-50, 5-8, and 3-7 results are identified for the three terms using the top-10, top-20 and top-50 method, respectively. Because the vector database can find sematic-related results by interpreting the contents using similarity ranking, it may introduce noises in the retrieval tasks, for example, the PRE is only 30% for the A-1 term using the top-10 method. Further increasing the $k$ value can find more correct entries, but the PRE value decreases significantly, indicating a dilution of correct information in the supplementary materials. The worst case is seen for the A-3 term, as the PRE is reduced by half considering a top-50 method instead.

Group B utilizes the exact phrases or sentences from the database as the search terms. Through manual interpretation, we can identify 23, 15, and 9 source chunks from the raw documents. Considering the relational database, we only find 1, 1, and 2 effective results for the B-1, B-2 and B-3 terms, respectively. This highlights the insufficiency of relational database when dealing with sematic search. In contrast, vector database outperforms relational database in this group. It provides almost all the information using the top-50 method, and the recall rate exceeds 95.7%. Further reducing the $k$ value to 20 and 10, the recall decreases from 60.9-100% to 55.6-75%. Taking a close examination on the results of top-10 method, it is found that the PRE is 90% and 100% for the B-1 and B-2 terms. This indicates that the amount of effective information exceeds the $k$ value, and a higher $k$ value is necessary to cover all the information for both cases. For the B-3 term, the PRE of the top-10 method is only 50%. We found that the returned irrelevant chunks of the B-3 term do not include the most critical keyword, i.e., electric tar catcher, and thus, similar chunks from other accident reports are identified through the semantic similarity of other words in the B-3 term. To avoid this issue, the similarity ranking should prioritize the name, time or location of the specific accident in our case. Otherwise, we shall expect a significant dilution of correct entities using the top-$k$ method when dealing with a retrieval including many words.

Group C mimics a test for natural language tasks. Relational database fails to return effective results for all three terms, leading to an insufficient knowledge supplement for LLMs. Vector database returns similar numbers of results for both the top-10 and top-50 methods, with the PRE-50 of the top-10 method ranging from 66.6-85.7%. Table A.1 in the Appendix A includes the top-10 results from vector database for the C-2 term, involving "Analysis the responsibility for the explosion accident at Shandong Sifang Installation Engineering Co., Ltd.". The returned text chunks show that six out of ten entries are highly relevant. The remaining entries vary in relevance; some are related to "responsibility analysis" without mentioning the company, and one entry is irrelevant that discuss the causes of other incidents.

The above analysis highlights the issue of relational database when dealing with queries of natural language (unstructured data). The vector database consistently outperforms the relational database, as vectors capture richer information using similarity ranking. Moreover, as the retrieval complexity increases from Group A to C, the PRE-50 of the top-10 method also increases, indicating that the majority of useful information may still be captured even when using a small $k$ value. Overall, we choose a $k$ value of 10 to achieve a practical trade-off between token consumption and the quality of supplementary knowledge provided.

### 3.2 LLM evaluation

In assessing the effectiveness of AI assistant tools such as ChatSOS, it is crucial that the evaluation criteria are in line with the needs of the users (Johri et al., 2023; Qu et al., 2024). Based on our research, we categorize the potential users into four groups from their practical requirements. As shown in Table 3, the general public, primarily nonspecialists, can use ChatSOS to access news or engage in scientific literacy. On the other hand, academic researchers, business personnel, and staff in national authority represent a spectrum of professionals with varying specific needs who demand more in-depth interactions with the tool, such as engaging in expert knowledge Q&A, accessing educational resources, coordinating inter-departmental collaboration guides, and drafting government documents.

**Table 3**
Potential users and practical applications

| Categories | Applications | Forms |
| --- | --- | --- |
| General public | News, popular science | Tweet, news report |
| Academic researchers | Scientific research, teaching | Q&A, research report, teaching material |
| Business personnel | Assess, training, emergency plan | Guidance, risk assessment report |
| Staff in national authority | Accident report, accident investigation | Official document, instruction |

Upon defining the fundamental user needs, a set of detailed evaluation criteria is proposed in Table 4. These criteria analyse the performance of AI assistants across multiple dimensions, with each assessment metric weighted according to its significance in practical applications. The final weighted scores provide a reference for the performance of AI assistants such as ChatSOS in real-world scenarios.

**Table 4**
Scoring criteria

| Categories | Weight |
| --- | --- |
| Relevance (whether it responds and is pertinent to the topic) | 20% |
| Accuracy (the correctness of the response and the facts presented are authentic) | 30% |
| Comprehensiveness (the thoroughness of the discussion on the topic) | 20% |
| Adaptability (the ability to tailor responses and suit specific tasks) | 15% |
| Clarity (expression organization, summarization skills, and elaboration) | 15% |

In accordance with the summary of specific needs and scoring criteria presented above, Table 5 presents the four tasks designed in this study that closely align with real-world scenarios: Task 1 involves explaining a technical terminology, primarily examining the direct application of general knowledge by the LLMs; Task 2 seeks advice within a specific task scenario, aiming to assess the model's understanding of the designated context and its utilization of general knowledge; Task 3 entails summarizing a specific accident, assessing the model capability to retrieve and integrate key information; Task 4 extends to a detailed analysis of an accident, examining analytical skills of LLM models. To conduct a thorough assessment of ChatSOS across different tasks, this study includes two widely used LLMs, i.e., ERNIE Bot and ChatGPT, for comparative analysis. The comparative test results are presented in Figures 6 to 9.

Table 5
Overview of the 4 tasks

|  | Task requirements |
|---|---|
| Task 1 | Technical terminology explanation |
| Task 2 | Advisory |
| Task 3 | Brief summary |
| Task 4 | Comprehensive analysis |

In Task 1, three LLMs are requested to elucidate the phenomenon of "detonation" and its influencing factors. The responses (Figure 6) comprise three sections, e.g., an overview, a detailed exposition of key points, and a conclusion. All models demonstrate good relevance to the topic, but vary significantly in their output information. ChatGPT offers a comprehensive overview emphasizing detonation as a chemical reaction characterized by rapid energy release, shockwaves, high temperatures, and significant pressure increases. This model is particularly beneficial for users seeking a general understanding without deep technical detail, as it elucidates the key factors and multifaceted nature of detonation. ERNIE Bot focuses on the mechanics of detonation as a specific type of combustion wave that propagates at supersonic speeds, particularly detailing how detonation occurs in confined spaces and the resultant substantial damage. This description is advantageous for those requiring insights into the physical processes of wave propagation and its effects in specific scenarios. Benefiting from the localized knowledge database, ChatSOS provides not only a general description of detonation but also incorporates practical safety considerations and crucial risk assessment information into its response, which is derived from practical applications. The performances of the three models are similar on this task, and their summary scores are closely matched, with values of 4.35, 4.20, and 5.00, respectively (Table 9).

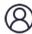

Fig. 6. Task 1 for the terminology explanation of "detonation". The bold sections only include the keywords instead, and the detailed information is included in the appendix.

In Task 2, all three models are inquired to provide advice on the self-rescue of operators in an explosion accident (Figure 7). ERNIE Bot is unable to return a valid answer due to relevant regulations, therefore, it has not been scored for this task. Both ChatGPT and ChatSOS structure their responses in an introduction-body-conclusion format. ChatGPT resolves the query from a general perspective, offering emergency actions that suit all accidents. Additionally, it includes irrelevant suggestions such as "Do Not Use Elevators", which are typically used for fire accidents in buildings. Furthermore, the suggested actions are presented in a random manner, lacking a systematic view of the proposed query. Based on the overall evaluation, we assign a score of 3.00 for accuracy and 2.00 for comprehensiveness. In contrast, ChatSOS gives an answer with high quality, it presumes the scenario involves an explosion accident occurring during operational procedures, and its advice is organized into six categories about action-oriented self-rescue measures. In the response, several professional recommendations, such as "activate emergency response procedures" and "use Personal Protective Equipment (PPE)", are included. Overall, the response of ChatSOS is specialized for explosion accidents, as requested, demonstrating a deep comprehension of the professional context and providing more effective recommendations for practical use.

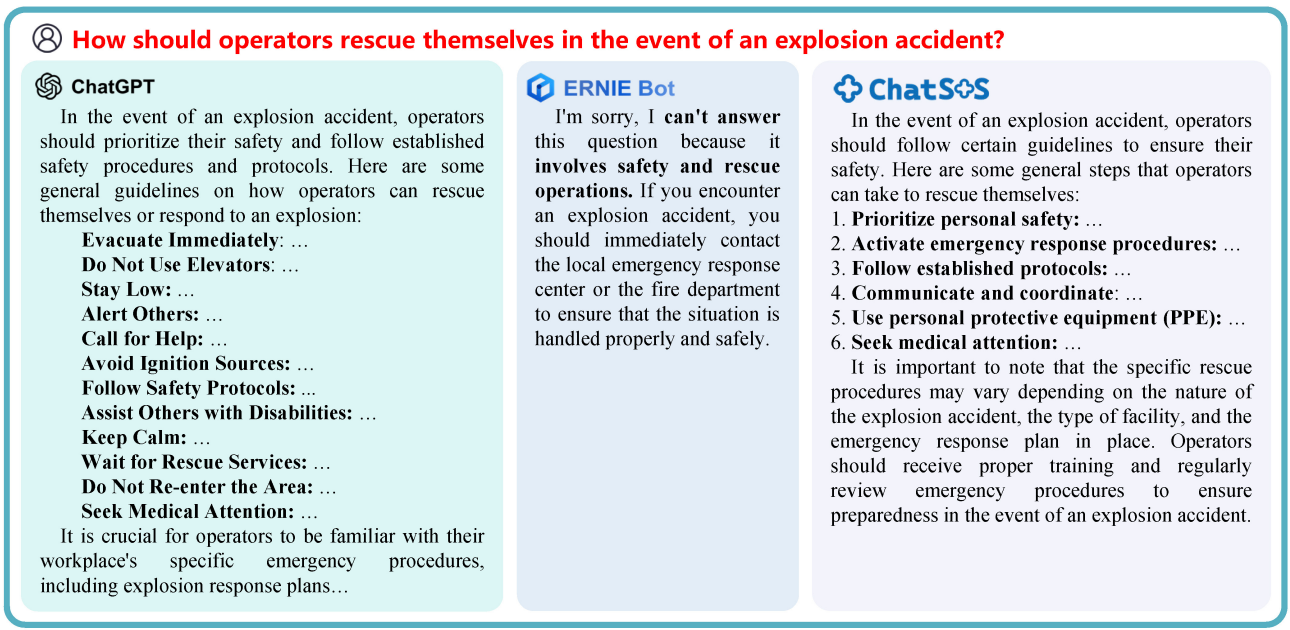

**Fig. 7.** Task 2 for seeking advice in a specific event. The bold sections only include the keywords instead, and the detailed information is included in the appendix.

In Task 3, three models are requested to summarize the "11·9 major explosion accident at Shanghai Qingpu Hengchang Logistics Co., Ltd.". The corresponding responses, as depicted in Figure 8, vary significantly in detail. ChatGPT does not provide a detailed summary but explicitly states that it lacks specific data about this event. The accident occurred on November 9, 2022, beyond the training corpus of ChatGPT up to January 2022, therefore, it has not been scored for this task. Despite this limitation, ChatGPT avoids model hallucination about the accident, preventing the generation of misleading contents. ERNIE Bot delivers a succinct summary that encapsulates the essential details of the accident, e.g., time, location, consequences, including casualties and economic losses, and causes. ERNIE Bot suggests that the cause of the accident is poor management and inadequate safety education. However, the above suggestion is incorrect, as the report indicates that the explosion resulted from the mishandling of a breached container that emitted flammable gases. ChatSOS provides an enhanced summary that includes key elements found in ERNIE Bot's response, optimized through preset prompt templates and enriched by detailed data from local database. It constructs a comprehensive narrative detailing the timeline of events leading to the explosion, focusing on both immediate impacts and root causes. ChatSOS integrates specific operational details with the broader consequences of the accident, offering a detailed overview of the event. In task 3, ChatSOS generally outperforms ERNIE Bot in terms of accuracy and comprehensiveness.

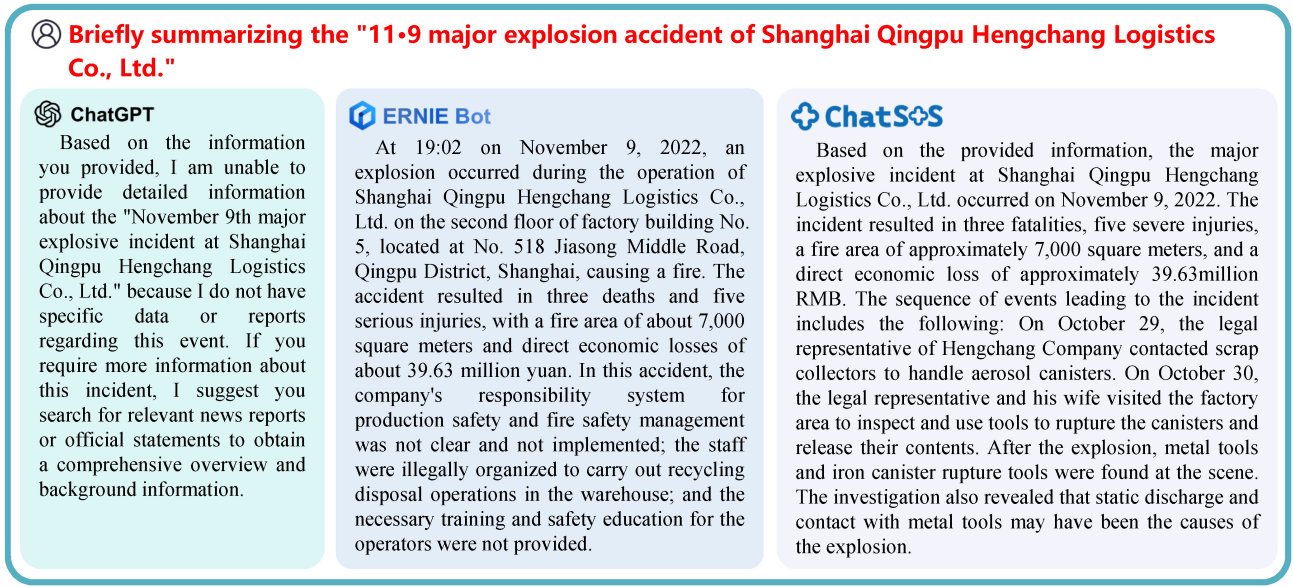

**Fig. 8.** Task 3 for summarizing an accident.

Figure 9 shows the results of the detailed analysis of the "Shandong Tianbao Chemical 10.21 Major Explosion Incident" by the three models. ChatGPT organizes its analysis into three sections: an accident overview, five analytical points, and a brief summary. The response covers five dimensions: accident background, details (including causes, impact range, rescue, and aftermath), impacts (casualties, property damage, and environmental impact), safety issues (chemical safety and government regulation), and lessons and prevention. The accident in question occurred within the timeframe covered by the training corpus of ChatGPT. However, the reply includes incorrect time and location as 2021/10/21 and Weifang City rather than the correct ones as 2015/10/21 and Pingyi. This is a typical example of model hallucinations. ERNIE Bot accurately captures the detailed characteristics of the requested accident, such as time, location, nature, and casualties. But, the response includes vague statements, such as "For some unknown reason, a violent chemical reaction occurred, ultimately leading to the explosion". This can be attributed to the insufficient information that is found by ERNIE Bot about the requested accident. Based on the evaluation, ChatGPT and ERNIE Bot get 1.00 and 3.00 points in accuracy, 3.00 and 2.00 points in comprehensiveness. ChatSOS produces a report that aligns closely with the original information in the database without errors. The use of a prompt template, which includes question-and-answer examples, allows ChatSOS to respond to the user's query from a reference framework, guiding the model to analyse the accident. The response is concise and focuses on the available information without introducing speculative content regarding missing regulatory and preventive measures in the accident context. In addressing complex issues related to accident details, ChatSOS still performs better than the other two models.

> **Complete Analysis of the "Shandong Tianbao Chemical 10.21 Major Explosion Incident"**
>
> **ChatGPT**
> The "Shandong Tianbao Chemical 10.21 Major Explosion Incident" refers to a significant chemical explosion accident that occurred in Linqiu County, **Weifang City**, Shandong Province, China, **on October 21, 2021**. Below is a comprehensive analysis of the incident:
> 1. Incident Background: The incident took place ...
> 2. Incident Details: The causes of chemical explosion accidents **typically include** equipment failures, chemical leaks.... A detailed investigation and analysis **are required to** determine the specific reasons...
> 3. Impact: The accident **may have resulted in** casualties... Chemical explosion accidents **often** involve chemical leaks that can lead to ... The explosion **may have caused significant damage** to factory ...
> 4. Safety Issues: Such incidents underscore safety issues in chemical plants...
> 5. Lessons and Prevention: After the accident, it is essential to learn from the incident...
>    In summary, the "Shandong Tianbao Chemical 10.21 Major Explosion Incident" was a significant explosion accident...
>
> **ERNIE Bot**
> **Title:** The October 21st Major Explosion Accident at Shandong Tianbao Chemical: Causes, Impacts, and Coping Strategies
>    On **October 21st, 2015**, a severe explosion accident occurred in Pingyi, Shandong, at the Tianbao Chemical enterprise. The accident caused **9 people to go missing** and **2 people to sustain injuries**...
>    **Firstly, let's take a look at** the causes of this accident... **For some unknown reason,** a violent chemical reaction occurred, ultimately leading to the explosion...
>    **Next, let's analyze** the impacts of this accident... had a significant impact on its **production and reputation**... to **repair equipment**, ... on the surrounding **environment** and **residents' health**. This requires the local government and relevant departments to take prompt measures for environmental monitoring and evacuation of residents.
>    **So, how should we respond to such a major accident?** Firstly, for Tianbao Chemical, **immediate production cessation**... **in-depth investigation** ... **actively cooperate with government** and ...
>    **Finally,** we should **recognize the importance** of industrial safety...
>
> **ChatSOS**
> Based on the provided information, we can provide the following comprehensive analysis of the Shandong Tianbao Chemical 10.21 Major Explosion Incident:
>    **Overview:** On **October 21, 2015**, at around **8:13 AM**... This incident resulted in...
>    **Responsibility Analysis:** Based on the investigation... categorized as a significant production safety liability accident. This implies that **the company bore responsibility** for production safety at the time of the accident.
>    **Cause of the Incident:** According to the given information, specific details have not been provided. Therefore, we cannot analyze the exact reasons for the Shandong Tianbao Chemical 10.21 Major Explosion Incident based on the available information.
>    **Impact:** ... resulted in **9 fatalities, 2 injuries,** and direct economic losses exceeding **19 million RMB**. Additionally, the incident **may have had some impact** on the surrounding environment and society, **but the specific details were not mentioned** in the provided information.
>    **Response and Recovery Measures:** According to the given information, there are no detailed specifics regarding the response and recovery measures taken in response to the Shandong Tianbao Chemical 10.21 Major Explosion Incident.
>    **Prevention and Mitigation Measures, Regulatory Compliance, Lessons Learned:** Based on the given information, ...**were not provided.**
> In conclusion, based on the given knowledge, we have conducted a comprehensive analysis of the Shandong Tianbao Chemical 10.21 Major Explosion Incident. However, due to the limited information provided, we cannot offer more detailed analysis and conclusions.

**Fig. 9.** Task 4 for analysing an accident in detail. The bold sections only include the keywords, and the detailed information is included in the appendix.

In Task 1, which involves explaining technical terminology, all three models perform adequately. For Task 2, a basic suggestion task, ERNIE Bot is occasionally hindered by policy restrictions, while ChatGPT's biases in predefined scenarios affect its accuracy and comprehensiveness. For tasks requesting the organization of information using LLM models, i.e., Tasks 3-4, ERNIE Bot's capability to perform online searches grants it an enhanced ability to gather information. Nevertheless, the model still occasionally exhibits hallucinations and faces limitations in knowledge integration and task comprehension. ChatGPT also exhibits serious model hallucinations in Task 4 due to training data limitations. Both ChatGPT and ERNIE Bot face difficulties in extracting and accurately presenting information from public corpora, failing to meet professional standards in handling sensitive or classified details.

In contrast, ChatSOS can provide professional and reliable outputs based on the prompt templates and a localized knowledge database with semantic retrieval technology. The comparative analysis of the four task scenarios suggests that the vector database-augmented LLM, i.e., ChatSOS, can significantly mitigate common issues such as insufficient training data and model hallucinations found in the baseline LLMs. The results from ChatSOS consistently demonstrate high reliability in terms of accuracy, professionalism, stability, and adaptability, demonstrating its outstanding performance in handling complex natural language tasks.

**Table 6**

Score summary[a]

|  |  | Rel. | Acc. | Comp. | Adapt. | Clar. | Sum |
|---|---|---|---|---|---|---|---|
|  | Weight | 20% | 30% | 20% | 15% | 15% | 100% |
| Task 1 | ChatGPT | 5.00 | 4.00 | 4.00 | 5.00 | 4.00 | 4.35 |
| | ERNIE Bot | 5.00 | 4.00 | 4.00 | 5.00 | 3.00 | 4.20 |
| | ChatSOS | 5.00 | 5.00 | 5.00 | 5.00 | 5.00 | 5.00 |
| Task 2 | ChatGPT | 4.00 | 3.00 | 2.00 | 4.00 | 5.00 | 3.45 |
| | ERNIE Bot | - | - | - | - | - | - |
| | ChatSOS | 5.00 | 5.00 | 5.00 | 5.00 | 5.00 | 5.00 |
| Task 3 | ChatGPT | - | - | - | - | - | - |
| | ERNIE Bot | 5.00 | 3.00 | 3.00 | 4.00 | 4.00 | 3.70 |
| | ChatSOS | 5.00 | 5.00 | 5.00 | 5.00 | 5.00 | 5.00 |
| Task 4 | ChatGPT | 5.00 | 1.00 | 3.00 | 5.00 | 4.00 | 3.25 |
| | ERNIE Bot | 5.00 | 3.00 | 2.00 | 4.00 | 5.00 | 3.65 |
| | ChatSOS | 5.00 | 5.00 | 5.00 | 5.00 | 5.00 | 5.00 |

[a] The score here refers to the relative performance of the three models. A score of 5 corresponds to the best performance among all three models, and the quality of the responses is expected from a junior graduate student with basic knowledge about safety engineering.

## 4. Conclusions

This study explores the application of LLMs within safety engineering, focusing on the development and implementation of LLMs. We introduce ChatSOS, an optimized strategy for building an AI-driven generative question-answering assistant with the integration of vector database and prompt engineering, into the realm of safety engineering, leveraging remarkable reliability in NLP tasks.

Our method involves constructing a database through text splitting and vector embedding, combines the vector database for semantic-based similarity searches with the relational database for data storage. The database evaluation demonstrates that using similarity search with the top-$k$ method achieves a higher quality of information, thus improving the supplemental knowledge provided to the LLMs. However, the top-$k$ method is still constrained in specific scenarios, especially when the retrieval terms contain more than one keyword. Future efforts will focus on refining keyword prioritization based on specific query contexts. Using the prompt engineering method, we provide specific instructions tailored to different tasks, with the database serving as an updatable knowledge source, and the LLMs could generate more professional and reliable outcomes. The prompt engineering in ChatSOS provides more convenient operations and controllable outcomes by updating the database only, unlike the fine-tuning method, which requires retraining the model each data update and demands substantial computational resources and datasets with strict standards. Performance comparisons among the three LLMs indicate that ChatSOS outperforms standard models such as ChatGPT in terms of reliability, accuracy, and comprehensiveness while also improving the adaptability and clarity of responses, presenting a novel and feasible solution for LLMs in addressing practical safety engineering problems.

Future work will further explore the potential of ChatSOS in practical applications by developing a more intelligent model agent equipped with specialized toolkits for complex scenarios, and by integrating advanced technologies such as knowledge graphs to augment its knowledge scope and optimize the retrieval process. We anticipate that these advancements will enable broader applications of ChatSOS in engineering practices and public education, thus providing more effective support for security management and accident prevention. Ultimately, ChatSOS is poised to make significant contributions to the advancement of safety engineering.

## Acknowledgements

This work was supported by the State Key Laboratory of Explosion Science and Technology (ZDKT21-01) and the Science and Technology Innovation Program of Beijing Institute of Technology (2022CX01028).

## Declaration of generative AI and AI-assisted technologies in the writing process

During the preparation of this work, the author(s) used ChatGPT and ERNIE Bot in order to generate the test results of LLMs in Section 3.2. After using this tool/service, the author(s) reviewed and edited the content as needed and take(s) full responsibility for the content of the publication.

# Appendix A.

## Table A.1

Top 10 document chunks retrieved by vector database for the term "C-2" in table 1.

| Score | Content | Source |
|---|---|---|
| 47.28 | III. Causes of the Accident 1. Direct Cause: Sifang Company's construction personnel conducted hot work near the central vent hole of the tank top while there was contaminated oil stored in the tank and the vent hole was open, which exceeded the scope permitted by the "Special Hot Work Safety Operation Permit", igniting the explosion gas mixture in the vapor space above the contaminated oil tank, resulting in explosion and combustion. 2. Indirect Causes: a. Sifang Company failed to fulfill the safety production main responsibility of the construction unit, and the management of the construction team was inadequate. (1) Confusion in safety management of Sifang Company. i.e. **Safety production responsibility system for all staff was not soundly established and implemented.** Specific safety production responsibility systems for special operation personnel were not formulated, and **the company's main person in charge, safety director, head of engineering management department, etc., failed to implement safety production responsibilities as required,** leading to inadequate implementation of hazard investigation and control, safety education and training, etc. | Investigation Report on the Major Explosion Accident at Shandong Sifang Installation Engineering Co., Ltd. in Dongying Port Economic Development Zone on "7·26" in 2022 |
| 49.49 | Investigation Report on the "7·26" Major Explosion Accident of Shandong Sifang Installation Engineering Co., Ltd. in Dongying Port Economic Development Zone, Shandong Province. On the morning of July 26, 2022, at around 9:30 a.m., a major explosion occurred during the construction process of the transformation of the contaminated oil tank of the carbon steel system in the VOCs comprehensive governance upgrade project of Dongying Sifang Installation Engineering Co., Ltd. (hereinafter referred to as "Sifang Company") at Dongying Haike Rui Lin Chemical Co., Ltd. (hereinafter referred to as "Rui Lin Company") in Dongying Port Economic Development Zone, Dongying City, resulting in 3 deaths, 2 injuries, and a direct economic loss of approximately 5.128 million yuan.<br>I. Overview of the Accident and Emergency Response<br>A. Sequence of Events Leading to the Accident | Investigation Report on the Major Explosion Accident at Shandong Sifang Installation Engineering Co., Ltd. in Dongying Port Economic Development Zone on "7·26" in 2022 |
| 55.34 | (3) Safety production responsibility system severely lacked implementation. **The company's hot work supervisors and managers did not diligently implement the safety production responsibility system for their positions, failed to effectively supervise the execution of special hot work operations, and did not promptly identify and stop construction personnel from operating beyond their authorized scope.** (4) Inadequate hazard identification and control. **The company's safety and production management personnel, as well as supervisors, did not conduct thorough hazard identification and control for risky operations related to the accident, failing to identify hazards that did not meet the conditions for hot work.** (5) Inadequate management of external construction teams. **The company's supervision of safety production conditions for external construction teams was lax, with inadequate daily management and training assessment.** Safety management personnel did not effectively conduct safety inspections and coordinate management, failing to promptly halt the Sifang Company's construction team from operating beyond the authorized scope of special hot work on the day of the incident.<br>3. **The Dongying Port Economic Development Zone Management Committee, along with its Economic Development Bureau and Emergency Management Bureau, failed to diligently fulfill their responsibilities for safety production supervision.** | Investigation Report on the Major Explosion Accident at Shandong Sifang Installation Engineering Co., Ltd. in Dongying Port Economic Development Zone on "7·26" in 2022 |
| 58.17 | ②Severe lack of implementation of safety production management system and operating procedures, especially the hazardous operation management system and operating procedures. The construction team did not diligently adhere to relevant laws, regulations, standards, and norms regarding hot work operations, as well as the company's provisions on "Fire Safety Management" and "Welder Safety Operating Procedures." Before the incident, on-site safety conditions were not ensured, and hot work operations were conducted on the tank top beyond the scope permitted by the "Special Hot Work Safety Operation Permit," where hazards such as non-emptying displacement and open facilities existed. ③ Inadequate management of external construction teams. The management of external construction teams was insufficient, as they were managed through subcontracting agreements, with only the transmission of the company's regulations and procedures through WeChat groups, without effective supervision. | Investigation Report on the Major Explosion Accident at Shandong Sifang Installation Engineering Co., Ltd. in Dongying Port Economic Development Zone on "7·26" in 2022 |
| 59.23 | ②Failure to diligently fulfill the engineering supervision responsibilities, as the cutting and hot work operation on the tank cage at the vent of the return air shaft was not detected by the supervisory personnel, who were absent from the site on the day of the accident.<br>(5) Xingda Blasting Company operated without obtaining the "Road Transport Operation Permit" and the "Civil Explosive Transportation Permit." Both drivers and escorts lacked professional qualifications. The company unlawfully transported civil explosive materials to Wucai Long Company using vehicles lacking hazardous goods transportation qualifications. Moreover, Xingda Blasting Company applied for the "Civil Explosive Transportation Permit" and input flow information into the "Shandong Civil Explosive Information System Network Service Platform" under its own name for the transportation of civil explosive materials to Wucai Long Company. Additionally, Xingda Blasting Company failed to complete relevant procedures for transporting civil explosive materials to Anda Civil Blasting Company's warehouse.<br>(6) Anda Civil Blasting Company | 2021 Shandong Wucailong Investment Co., Ltd. Huashan Gold Mine in Qixia City "1·10" |

**Table A.1**

Top 10 document chunks retrieved by vector database for the term "C-2" in table 1. (Continued)

| Score | Content | Source |
|---|---|---|
| 59.47 | Analyzing factors such as gas pipeline leakage, formation of confined spaces, ignition of gas sources, and hazard identification and control, the indirect causes of this accident are as follows: Firstly, the violation of regulations in construction led to the occurrence of hidden dangers. In March 2005, Dongfeng Gas Company laid the involved pipeline (D57×4) without the approval of the competent authority. At that time, the pipeline had not yet passed under the relevant buildings. In October 2008, Dongfeng Gas Company unlawfully carried out partial modifications to the involved pipeline's medium-pressure branch. After the modification, the pipeline passed through the confined space beneath the relevant buildings, creating a safety hazard. | Investigation Report on the Major Gas Explosion Accident at Yanhu Community Farmers' Market in Zhangwan District, Shiyan City, Hubei Province, on "6·13" in 2021 |
| 59.80 | Company employee Sun immediately called Zhang to inform him of the situation. Zhang then went to the Victory Petroleum Administration's Coastal Hospital and Xianhe Funeral Parlor to confirm the identities of the deceased. Subsequently, Zhang reported the accident to Ma by phone. Ma later reported the accident to Sifang Company's legal representative, Zhang Mosen, and vice president Liu. However, **Sifang Company failed to report the accident to the emergency management department at the accident site and relevant departments responsible for safety supervision and management as required by regulations, constituting a concealment of the accident.** | Investigation Report on the Major Explosion Accident at Shandong Sifang Installation Engineering Co., Ltd. in Dongying Port Economic Development Zone on "7·26" in 2022 |
| 60.57 | 6. Inadequate safety education. On-site workers had a weak safety awareness, poor self-protection capabilities, and lacked awareness of the dangers of unsealing, sealing, and ventilating to discharge gas. There were instances of unauthorized commands and operations on-site.<br>(3) Nature of the Accident: Based on on-site surveys, investigation evidence, and technical analysis, the accident investigation team determined that the incident was a liability accident. | Investigation Report on the Major Gas Explosion Accident at Jinan Li Fu Coal Mine Co., Ltd. on "12.15" in 2014 |
| 61.78 | (1) Illegal construction: The company commenced construction without obtaining approvals from relevant departments such as land, planning, housing construction, safety supervision, fire control, and environmental protection. Despite being lawfully ordered to halt construction by environmental, safety supervision, and housing construction departments, the company evaded supervision, refused to comply with the cessation of construction orders, and continued construction without authorization.<br>(2) Improper feeding and commissioning: The company failed to strictly adhere to the "Safety Commissioning Regulations for Chemical Plant Equipment in Shandong Province" by conducting the "three inspections and four identifications" on the accident-prone equipment. It did not organize reviews of commissioning plans and safety conditions, establish commissioning management organizations, and conducted commissioning activities concurrently with construction without proper authorization. Construction personnel were housed irregularly in the commissioning plant area, and preparatory commissioning work such as pressure testing, blowdown, air tightness testing, individual equipment commissioning, and instrument calibration were not carried out in strict accordance with relevant regulations. | Investigation Report on the Major Explosion Accident at Shandong Binyuan Chemical Co., Ltd. in Dongying City, Shandong Province, on "8·31" in 2015 |
| 61.80 | At 7:44 a.m., two workers from Sifang Company climbed onto the top of the contaminated oil tank and began transferring tools onto the tank top. At 8:30 a.m., Liu handed over duties to the next shift leader, Zhang, while Che handed over duties to the next operator and hot work supervisor, Wang. The handover involved Sifang Company's prefabrication work in the blue barricaded area of the water tank zone. Around 8:46 a.m., Sifang Company workers started dismantling pipe sections on the tank top, completing the dismantling around 8:54 a.m. At approximately 8:58 a.m., Wang arrived at the prefabrication work site to supervise the operation. Around 9:23 a.m., three Sifang Company workers began unauthorized welding work near the newly installed valve bracket on the tank top. At 9:30:37 a.m., the contaminated oil tank exploded, causing the tank body to rupture, resulting in the death of one Sifang Company worker, the ineffective rescue of two workers who were sent to the hospital, and minor injuries to two Rui Lin Company personnel involved in the rescue operation.<br>(2) Emergency Response Situation: | Investigation Report on the Major Explosion Accident at Shandong Sifang Installation Engineering Co., Ltd. in Dongying Port Economic Development Zone on "7·26" in 2022 |

**Appendix B:** Complete Answer for Task 1 in Section 3.1

**Human:** Please explain the detonation phenomenon and its factors.

**ChatGPT:**
The detonation phenomenon refers to a rapid and intense form of chemical reaction, often associated with the explosion of materials like explosives or volatile gases. During detonation, a combination of factors leads to an almost instantaneous release of a large amount of energy, creating a shockwave, high temperatures, and a rapid increase in pressure. Here are the key factors that contribute to detonation:
Chemical Composition: Detonation typically involves materials with high energy content, such as explosives or reactive chemicals. These substances contain chemical bonds that can rapidly release energy when they break.
Oxygen or Oxidizer: A source of oxygen or an oxidizing agent is required to support the combustion reactions in detonation. Oxygen is a common oxidizer, but other chemicals can also serve this role.
Initiation: Detonation requires an initial energy input to start the reaction. This can be provided by a spark, heat, impact, or other forms of energy. The initiation step is critical to overcoming the activation energy barrier for the explosive reaction.
Confinement: Detonation often occurs in a confined space, where the expanding gases from the reaction encounter resistance from the surroundings. This confinement increases the pressure, leading to a more powerful explosion.
High Reaction Rates: Detonation reactions have extremely high reaction rates. This means that once initiated, the reaction proceeds very quickly, often at or near the speed of sound.
Shockwave Formation: The rapid release of energy in detonation generates a shockwave that travels outward from the explosion point. This shockwave can cause significant damage and is responsible for the characteristic destructive nature of detonations.
Overall, the detonation phenomenon is a complex interplay of chemical reactions, energy release, and physical conditions that result in a rapid and violent explosion. Understanding these factors is crucial for safety and risk management when dealing with explosives or other potentially explosive materials.

**ERNIE bot:**
Detonation is a type of combustion wave that propagates at supersonic speeds. When a mixture of fuel and air in a confined space is ignited, the resulting combustion wave compresses the surrounding gas mixture, causing it to explode. This rapid and violent explosion produces a strong shock wave that propagates through the confined space at supersonic speeds, often resulting in significant damage to nearby structures or objects.
The factors that affect detonation include:
Pressure: High pressure environments can increase the likelihood of detonation. As the pressure in a confined space increases, the mixture of fuel and air becomes more compressed, making it easier to reach the ignition temperature and propagate a combustion wave.

Temperature: High temperatures can also increase the likelihood of detonation. As the temperature of a confined space increases, the mixture of fuel and air becomes more volatile, increasing its tendency to ignite and propagate a combustion wave.

Confinement: The confined nature of the space can also contribute to detonation. As the mixture of fuel and air is confined within a small space, it has nowhere to expand, which can lead to a rapid increase in pressure and temperature, making it easier to reach the ignition temperature and propagate a combustion wave.

Chemical properties: The chemical properties of the fuel and air mixture can also affect detonation. Some fuels are more volatile than others, and certain mixtures of fuel and air can be more likely to ignite and propagate a combustion wave.

In summary, detonation is a violent explosion that occurs when a mixture of fuel and air in a confined space is ignited. The factors that affect detonation include pressure, temperature, confinement, and chemical properties.

**ChatSOS:**
Detonation is a rapid and violent combustion process that occurs when a high-energy explosive material undergoes a self-sustaining chemical reaction. It is characterized by a shock wave that travels at supersonic speeds through the explosive material, causing a sudden release of energy.

There are several factors that can contribute to the detonation phenomenon:

1. Sensitivity: The sensitivity of an explosive material refers to its ability to undergo detonation when subjected to an initiating stimulus, such as heat, impact, or friction. Highly sensitive materials have a lower threshold for detonation and are more likely to undergo this rapid combustion process.

2. Confinement: The confinement of an explosive material plays a crucial role in the detonation process. When the explosive is confined, such as in a container or a pipeline, the shock wave generated by the initial explosion can be reflected back into the explosive, leading to a self-sustaining detonation.

3. Composition: The composition of the explosive material also affects its detonation characteristics. Different combinations of chemicals can result in varying levels of stability and sensitivity. Factors such as the size and distribution of particles, the presence of impurities, and the presence of oxygen-supplying compounds can all influence the likelihood of detonation.

4. Initiation: The initiation of detonation requires an initiating stimulus, commonly known as a detonator or a blasting cap. This stimulus provides the necessary energy to start the rapid combustion process. The type and strength of the detonator used can determine the likelihood and intensity of detonation.

It is important to note that detonation is a complex phenomenon, and understanding its factors is crucial for the safe handling and storage of explosive materials. Proper storage, handling procedures, and risk assessment are necessary to mitigate the potential hazards associated with detonation.

**Appendix C:** Complete Answer for Task 2 in Section 3.1

**Human:** How should operators rescue themselves in the event of an explosion accident?

**ChatGPT:**
In the event of an explosion accident, operators should prioritize their safety and follow established safety procedures and protocols. Here are some general guidelines on how operators can rescue themselves or respond to an explosion:
Evacuate Immediately: If there is any indication or suspicion of an explosion, operators should evacuate the area as quickly as possible. Follow the designated evacuation routes and assembly points, if available.
Do Not Use Elevators: Avoid using elevators during an emergency, as they may become inoperative or unsafe. Use stairwells and exit staircases instead.
Stay Low: If smoke or noxious gases are present, stay close to the floor where the air is likely to be less contaminated. Use a cloth or garment to cover your nose and mouth to reduce inhalation of harmful substances.
Alert Others: If you encounter other individuals while evacuating, inform them of the situation and guide them to safety if possible.
Call for Help: Use your phone or a nearby emergency phone to call for help and inform emergency services of the explosion.
Avoid Ignition Sources: Do not use lighters, matches, or any other potential ignition sources in the vicinity of a suspected explosion, especially if there is a possibility of gas leaks.
Follow Safety Protocols: If you are trained in handling hazardous materials or working in an environment prone to explosions, follow the safety protocols and procedures established for your specific workplace or industry.
Assist Others with Disabilities: If you encounter individuals with disabilities or mobility issues, offer assistance in a safe and efficient manner.
Keep Calm: It's essential to remain as calm as possible during an emergency. Panic can lead to poor decision-making and hinder your ability to respond effectively.
Wait for Rescue Services: In some cases, it may not be safe to evacuate the area immediately. If you cannot escape, find a safe location to take cover and wait for rescue services to arrive.
Do Not Re-enter the Area: After evacuating, do not re-enter the affected area until it has been declared safe by emergency responders or designated authorities.
Seek Medical Attention: If you or anyone else is injured during the explosion, seek medical attention promptly.
It is crucial for operators to be familiar with their workplace's specific emergency procedures, including explosion response plans, and to participate in regular safety drills and training to ensure a swift and effective response in the event of an explosion or any other emergency. Additionally, operators should always prioritize their safety and the safety of others during such accidents.

**ERNIE bot:**

I'm sorry, I can't answer this question because it involves safety and rescue operations. If you encounter an explosion accident, you should immediately contact the local emergency response center or the fire department to ensure that the situation is handled properly and safely.

**ChatSOS:**

In the event of an explosion accident, operators should follow certain guidelines to ensure their safety. Here are some general steps that operators can take to rescue themselves:

1. Prioritize personal safety: The first and foremost step is to ensure your own safety. Operators should immediately move to a safe location away from the explosion site. This may involve evacuating the area or seeking shelter in a designated safe zone.
2. Activate emergency response procedures: Operators should be familiar with the emergency response procedures specific to their facility. This may include activating alarms, notifying emergency services, and initiating evacuation protocols.
3. Follow established protocols: Operators should follow established protocols for handling explosion accidents. This may involve shutting down equipment, isolating affected areas, or implementing containment measures to prevent further damage or spread of hazardous materials.
4. Communicate and coordinate: Operators should communicate with other team members and emergency responders to share information and coordinate rescue efforts. This includes providing updates on the situation, sharing knowledge of the facility and equipment, and following instructions from emergency personnel.
5. Use personal protective equipment (PPE): Operators should utilize appropriate personal protective equipment to protect themselves from potential hazards, such as explosion debris, toxic gases, or fires. This may include wearing safety goggles, helmets, gloves, and respiratory protection, as required.
6. Seek medical attention: After ensuring their own safety, operators should seek medical attention if they have sustained any injuries or if there is a possibility of exposure to hazardous materials. It is important to promptly report any injuries or symptoms to medical professionals for proper evaluation and treatment.

It is important to note that the specific rescue procedures may vary depending on the nature of the explosion accident, the type of facility, and the emergency response plan in place. Operators should receive proper training and regularly review emergency procedures to ensure preparedness in the event of an explosion accident.

**Appendix D:** Complete Answer for Task 4 in Section 3.1

**Human:** Complete Analysis of the "Shandong Tianbao Chemical 10.21 Major Explosion Accident".

**ChatGPT:**

The "Shandong Tianbao Chemical 10.21 Major Explosion Incident" refers to a significant chemical explosion accident that occurred in Linqiu County, Weifang City, Shandong Province, China, on October 21, 2021. Below is a comprehensive analysis of the incident:

**1. Incident Background:**
Location: Tianbao Chemical Co., Ltd. in Linqiu County, Weifang City, Shandong Province
Date: October 21, 2021.
Type: A major explosion accident leads to significant casualties and property damage.

**2. Incident Details:**
Cause: The causes of chemical explosion accidents typically include equipment failures, chemical leaks, operational errors, ignition sources, and various other factors. A detailed investigation and analysis are required to determine the specific reasons.
Impact area: The explosion caused damage to nearby buildings and factories, impacting the surrounding community and possibly resulting in casualties and missing persons.
Rescue: After the accident, emergency rescue personnel promptly arrived the scene to conduct rescue operations and firefighting. Injured individuals were transported to hospitals for treatment.
Subsequent treatment: After the incident, local government and relevant authorities need to conduct a thorough investigation to identify the causes, and ensure that similar incidents are prevented in the future, and provide support to the victims.

**3. Impact:**
Casualty: The accident may have resulted in casualties, including fatalities and injuries.
Environmental impact: Chemical explosion accidents often involve chemical leaks that can lead to environmental pollution, posing risks to ecosystems and water sources in the vicinity.
Property loss: The explosion may have caused significant damage to factory equipment and buildings, resulting in substantial property losses.

**4. Safety Issues:**
Chemical safety: Such incidents underscore safety issues in chemical plants, including equipment maintenance, employee training, and chemical management.
Government supervision: The incident also raises concerns about government oversight, emphasizing the need for regulatory authorities to ensure strict compliance with safety regulations in chemical enterprises.

**5. Lessons and Prevention:**
Lessons: After the accident, it is essential to learn from the incident, conduct a thorough investigation into its causes, improve chemical production processes, and enhance safety management in chemical enterprises.
Prevention: Through rigorous safety oversight, standardized operating procedures, and training, similar accidents can be minimized.

In summary, the "Shandong Tianbao Chemical 10.21 Major Explosion Incident" was a significant explosion accident involving a chemical factory that resulted in severe casualties and property damage. Such incidents highlight the importance of industrial safety and the need to take measures to prevent the recurrence of similar events and ensure the safety of industrial production processes.

**ERNIE bot:**
Title: The October 21st Major Explosion Accident at Shandong Tianbao Chemical: Causes, Impacts, and Treatment Strategies

On October 21st, 2015, a severe explosion accident occurred in Pingyi, Shandong, at the Tianbao Chemical enterprise. The accident caused 9 missing persons and two injuries. For such an accident, we must conduct an in-depth analysis of its causes, impacts, and treatment strategies.

Firstly, let's take a look at the causes of this accident. According to reports, Tianbao Chemical was conducting production operations at the time of the accident. For some unknown reason, a violent chemical reaction occurred, ultimately leading to an explosion. Specifically, the cause of the accident may be due to equipment failure, operational errors, inadequate safety management, or other factors. As the investigation has not yet reached a conclusion, the exact cause cannot be determined, but it is certain that this was a serious industrial safety accident.

Next, let's analyze the impacts of this accident. Firstly, for Tianbao Chemical, this accident had a significant impact on its production and reputation. The company may need to invest a significant amount of time and resources to repair equipment, resume production, and handle related compensations. In addition, the public trust for Tianbao Chemical may decrease, posing pressures on its future operations and development. Secondly, for the local community, the explosion shockwave and harmful substances may impact the surrounding environment and residents' health. This requires the local government to take prompt measures for environmental monitoring and evacuation of residents.

So, how should we respond to such a major accident? Firstly, for Tianbao Chemical, immediate production cessation is necessary, along with an in-depth investigation of the accident's causes and effective measures to prevent similar accidents. At the same time, the company should cooperate with government by providing necessary assistance. Secondly, for the local government, they should activate emergency plans, organize rescue forces for on-site response and rescue operations. Simultaneously, they should strengthen environmental monitoring and evacuation efforts to ensure public safety. Additionally, the government and relevant departments should also strengthen post-accident handling efforts, including compensation for victims and their families, as well as punishment for the company.

Finally, we should recognize the importance of industrial safety. Both enterprises and government and relevant departments should strengthen safety supervision and risk prevention efforts to ensure the safety of people's lives and property. Only in this way can we truly achieve social harmony and stability as well as sustainable development.

**ChatSOS:**
Based on the provided information, we can provide the following comprehensive analysis of the Shandong Tianbao Chemical 10.21 Major Explosion Incident:

**Overview:** On October 21, 2015, at ~8:13 am, an explosion occurred in the manual detonator production workshop of the Shandong Tianbao Chemical Co., Ltd., located in the Nanshan production area of Pingyi County, Linyi City, Shandong Province. This incident resulted in 9 fatalities, 2 injuries, and direct economic losses exceeding 19 million RMB.

**Responsibility Analysis:** Based on the investigation, the Shandong Tianbao Chemical 10.21 Major Explosion Incident was categorized as a significant production accident. This implies that the company should be responsible for production safety at the time of the accident.

Cause of the Incident: According to the given information, specific details about the causes of the accident have not been provided. Therefore, we cannot analyze the exact reasons based on the available information.

**Impact:** The Shandong Tianbao Chemical 10.21 Major Explosion Incident resulted in 9 fatalities, 2 injuries, and direct economic losses exceeding 19 million RMB. Additionally, the incident may have some impact on the surrounding environment and society, but the specific details of these effects were not mentioned in the provided information.

**Response and Recovery Measures:** According to the given information, there are no detailed specifics regarding the response and recovery measures taken in response to the Shandong Tianbao Chemical 10.21 Major Explosion Incident.

**Regulatory Compliance:** The provided information does not mention whether the incident complied with relevant regulatory requirements and compliance standards.

Prevention and Mitigation Measures: Due to the limited information provided, we are unable to analyze the prevention and mitigation measures related to the Shandong Tianbao Chemical 10.21 Major Explosion Incident.

**Lessons Learned:** Based on the given information, specific lessons learned from the Shandong Tianbao Chemical 10.21 Major Explosion Incident were not provided.

In conclusion, based on the given knowledge, we have conducted a comprehensive analysis of the Shandong Tianbao Chemical 10.21 Major Explosion Incident. However, due to the limited information provided, we cannot offer more detailed analysis and conclusions.